**Title:** Application and Validation of Geospatial Foundation Model Data for the Prediction of Health Facility Programmatic Outputs - A Case Study in Malawi

**Authors:** Lynn Metz[1], Rachel Haggard[1], Michael Moszczynski[1], Samer Asbah[1], Chris Mwase[1], Patricia Khomani[1], Tyler Smith[1], Hannah Cooper[1], Annie Mwale[2], Arbaaz Muslim[3], Gautam Prasad[3], Mimi Sun[3], Tomer Shekel[3], Joydeep Paul[3], Anna Carter[3], Shravya Shetty[3], Dylan Green[1, *]

**Affiliation:** 1 Cooper/Smith, 2 Government of Malawi, 3 Google Research

**Correspondence:** dylan@coopersmith.org

**Word Count:** 3,010

**Tables/Figures:** 2/2

**Funding:** Google.org

# Abstract (286)

The reliability of routine health data in low and middle-income countries (LMICs) is often constrained by reporting delays and incomplete coverage, necessitating the exploration of novel data sources and analytics. Geospatial Foundation Models (GeoFMs) offer a promising avenue by synthesizing diverse spatial, temporal, and behavioral data into mathematical embeddings that can be efficiently used for downstream prediction tasks. This study evaluated the predictive performance of three GeoFM embedding sources - Google Population Dynamics Foundation Model (PDFM), Google AlphaEarth (derived from satellite imagery), and mobile phone call detail records (CDR) - for modeling 15 routine health programmatic outputs in Malawi, and compared their utility to traditional geospatial interpolation methods. We used XGBoost models on data from 552 health catchment areas (January 2021–May 2023), assessing performance with the coefficient of determination ($R^2$), and using an 80/20 training and test data split with 5-fold cross-validation used in training. While predictive performance was mixed and varied by indicator, the embedding-based approaches improved upon baseline geostatistical methods in 13 of 15 (87%) indicators tested. A Multi-GeoFM model integrating all three embedding sources produced the most robust predictions, achieving average 5-fold cross validated $R^2$ values for indicators like population density (0.63), new HIV cases (0.57), and child vaccinations (0.47) and test set $R^2$ of 0.64, 0.68, and 0.55, respectively. Prediction was poor for prediction targets with low primary data availability, such as TB and malnutrition cases. These results demonstrate that GeoFM embeddings imbue a modest but meaningful predictive improvement for select health and demographic outcomes in an LMIC context. We conclude that the strategic integration of multiple GeoFM sources via Multi-GeoFM modeling is an efficient and valuable tool for supplementing and strengthening constrained routine health information systems.

## Background

Reliable and timely health data are crucial for effective health system monitoring, resource allocation, and policy planning (1, 2). However, current data systems often face challenges such as reporting delays, incomplete coverage, and variable data quality (3, 4). As a result, researchers have begun turning to novel and nontraditional data sources to supplement routine reporting systems (5, 6). For example, geospatial data can be used to provide insights into population, environment, and infrastructure, and changes in these factors over time (7). Mobile phone data can capture patterns in population movement and behavior to help predict disease transmission and health care access (8). Internet search data can contribute to infectious disease surveillance by identifying trending health concerns and regional patterns (9).

Advances in artificial intelligence, specifically neural networks and geospatial foundational models (GeoFMs), have expanded the potential uses of these diverse data sources (10). GeoFMs are machine learning models that synthesize large amounts of geospatial-relevant data into embeddings or mathematical representations that capture complex spatial, temporal, and behavioral patterns. These embeddings can then be used in downstream tasks to predict outcomes in areas such as health, environment, and economics, while requiring only a fraction of the compute power and resources typically needed to work with large geospatial datasets (11, 12). Social media has also emerged as an interesting data source for understanding behaviors, reactions, or information that is indicative of disease outbreaks (5).

The successful application of GeoFM embeddings in downstream tasks has been documented, specifically in high-income settings (12). Google successfully utilized its population dynamics foundational model (PDFM) embeddings to predict health indicators, socioeconomic factors, and environmental measurements in the United States (12). Similarly, Satellite Contrastive Location-Image Pretraining (SatCLIP) embeddings have been shown to improve performance in multiple geospatial tasks such as population density estimation, temperature prediction, and species recognition (13). Other GeoFM models including Prithvi, Presto, Clay, TESSERA, and Galileo, have motivated and laid the groundwork for further innovation in this technical area (14, 15, 16, 17, 18). Additionally, knowledge graph embeddings generated on intensive care (ICU) data improved predictive performance of ICU readmissions (19). Despite the successes, a gap in literature exists when it comes to the application of GeoFM embeddings in low and middle income countries (LMICs).

The benefit of these GeoFM embeddings may be particularly impactful in LMICs, where resource constraints and connectivity challenges contribute to data system limitations (20). For example, in Malawi, approximately 60% of the population report access to a mobile phone, but only 15% report owning a smartphone (up to 30% in cities), and overall internet penetration is also 15% (21). These limitations underscore the importance of evaluating how well mobile and internet-derived data sources, as well as geospatial embeddings, can support health system monitoring in low-connectivity settings. At the same time, Malawi faces a dual burden of disease: communicable, maternal, perinatal, and nutritional conditions account for 51% of deaths, compared to only 16% in the United States, while noncommunicable diseases (NCDs) contribute to 40% of deaths in Malawi versus 77% in the United States (22, 23). This epidemiologic profile differs substantially from high-income countries, where NCDs dominate, making Malawi a particularly relevant setting to test whether novel data sources and embeddings can capture health dynamics across both infectious and chronic disease burdens. Finally, Malawi is among many countries affected by reductions in global development assistance for health, which supports core data systems leading to a decline in health data quality and availability (24).

To address the existing knowledge gap, we evaluated and compared the predictive performance of three sets of GeoFM embeddings for modeling health indicators across Malawi, with the objective of identifying if these embeddings provide predictive value beyond traditional geospatial modeling approaches in an LMIC setting; and if so which data sources and embedding strategies most effectively enhance predictive accuracy.

## Methods

**Data Sources**
*Administrative Shapefiles*
Geospatial data on health catchment area boundaries, including catchment population estimates, were obtained from the Government of Malawi. A total of 552 catchments were included in our analysis.

*GeoFM Sources*
Three GeoFM sources were utilized, Google PDFM, Google AlphaEarth, and mobile phone call detail records metadata (CDR) (Table 1).

**Table 1:** Characteristics of GeoFM Sources

| Source | Data | Years | Definition |
|---|---|---|---|
| Google PDFM* (12) | Aggregated search trends | Jan - Jul 2024 | 16-dimensional embeddings, at the catchment level, generated from Google Search data, Google Maps data, "busyness" indicators, and weather and air quality data. |
| | Maps | Apr 2025 | |
| | Busyness | Jul 2023 | |
| | Weather and air quality | Jul 2023 | |
| Google AlphaEarth (25) | SENTINEL-2 | 2023 | 64-dimensional embeddings, at a 10m pixel resolution, generated from multiple satellite imagery sources and land use metrics. |
| | LANDSAT 8/9 | | |
| | SENTINEL-1 | | |
| | PALSAR-2 | | |
| | ERA5-LAND | | |
| | GEDI | | |
| | GRACE | | |
| | GLO-30 | | |
| | NLCD | | |
| | Land use evaluation metrics | | |
| CDR (26) | Mobile phone call and SMS metadata | Mar 2020 - Sep 2021 | 10-dimensional embeddings, at the catchment level, derived from mobile phone metadata including anonymized subscriber counts labeled spatiotemporally. |

*N.B. These catchment level Malawi embeddings were created by Google Research specifically for this research project.

*Downstream Prediction Target Data*

Prediction targets were derived from the District Health Information System (DHIS2) and the Malawi National Laboratory Information Management System (LIMS) for the period Jan 2021 - May 2023. DHIS2 is an open-source health management information system widely used by ministries of health to collect, manage, and analyze routine health facility-based data. LIMS is a national database used to manage laboratory testing and results. Health facility latitude and longitude coordinates were used to assign facilities to catchment areas through spatial joins with catchment geospatial polygons. Data were aggregated by taking the weighted mean of all facility level data within the catchment. To improve data quality, for indicators reported on a monthly basis, facilities with missing values for more than 23 of the 29 study months were excluded from analysis. For indicators reported quarterly, facilities with missing values for more than 7 of 10 quarters were excluded. In addition, facilities with ≥75% of values reported as zero were excluded, under the assumption that these zeros primarily reflect data reporting gaps rather than true absence of events. All indicators are defined in Table 2. Non-rate indicators were log-transformed prior to model inclusion to normalize distributions and then inverse-transformed before model evaluation.

**Table 2:** Characteristics of Downstream Prediction Targets

| Source | Data | Years | N | Numerator | Denominator |
|---|---|---|---|---|---|
| Catchment data | Population density | Mar 2020 - Sep 2021 | 503 | The number of people in each catchment area. | The area in kilometers$^2$ of each catchment. |
| DHIS 2 | HIV test positivity rate | Mar 2020 - Sep 2021 | 434 | The number of new HIV cases. | The number of HIV tests performed in the catchment. |
| DHIS 2 | Malaria case rate | Mar 2020 - Sep 2021 | 450 | The number of new Malaria cases. | The catchment population. |
| DHIS 2 | Antenatal care access | Mar 2020 - Sep 2021 | 463 | The number of pregnant women to go to at least four antenatal care visits. | The total number of pregnant women in the catchment. |
| DHIS 2 | TB case rate | Mar 2020 - Sep 2021 | 327 | The number of new TB cases detected and put on treatment. | The catchment population. |
| DHIS 2 | STI cases | Mar 2020 - Sep 2021 | 459 | The number of new STI diagnoses. | NA |
| DHIS 2 | Malnutrition cases | Mar 2020 - Sep 2021 | 227 | The number of children classified as malnourished. | NA |
| DHIS 2 | Child vaccinations | Mar 2020 - Sep 2021 | 452 | The number of children fully vaccinated. | NA |
| DHIS 2 | HIV diagnosis at first antenatal care visit | Mar 2020 - Sep 2021 | 463 | The number of pregnant women who were tested positive for HIV at their first antenatal care visit. | The number of pregnant women who were tested for HIV at their first antenatal care visit. |
| LIMS | Unsuppressed Viral Load (HIV) rate | Mar 2020 - Sep 2021 | 373 | The number of people living with HIV who had an unsuppressed viral load result. | The number of viral load tests run in the catchment. |

**Statistical Analysis**

All analyses were performed using Python. Data were split randomly into 80% training and 20% testing sets with a 5-fold cross-validation approach applied to the training set, while each prediction target was evaluated individually on the test set.

Two traditional spatial prediction approaches were implemented as baselines. The first method implemented was Inverse Distance Weighting (IDW) with a power parameter of 2, using the six nearest neighbors at each prediction point. The second method was ordinary kriging using a spherical variogram model. For both methods the geographic inputs were the centroid latitude/longitude of each catchment area.

We implemented XGBoost models to predict each prediction target. Hyper-parameter tuning was performed using a grid search over the following parameters: learning rate [0.01, 0.05, 0.1, 0.3], maximum tree depth [2, 3, 4, 6, 8], and number of boosting rounds [50, 100, 200, 400]. The best performing parameter set was identified separately for each combination of embedding source and prediction target during cross-validation. Separate models were trained for each combination of prediction target and embedding source. In addition, a Multi-GeoFM model was trained using all embedding features combined to evaluate the predictive value of integrating multiple embedding-based models.

Model performance was assessed using the coefficient of determination ($R^2$), which is reported for all prediction targets and modeling approaches and as an average of the 5-fold cross-validation process as well as the $R^2$ from the test set. The standard deviation of $R^2$ is calculated from the 5-fold cross-validation splits and provided as a measure of model stability and variability in predictive performance. For XGBoost models, the $R^2$ standard deviation reflects variation across folds in the cross-validation procedure. For the deterministic methods (IDW and kriging), which do not include random model components, the reported $R^2$ standard deviation represents variation across spatial folds rather than random model variability. This approach ensured consistent evaluation and comparison of predictive accuracy across embedding sources and the Multi-GeoFM model.

## Results

Figure 1 summarizes the $R^2$ values of all predictive models, including the baseline methods and the embedding XGBoost models for rate-type prediction targets. Figure 2 presents the same metrics in the same fashion, but for count-type prediction targets. For both types of indicators, performance differed widely based on the model type, embedding source utilized, and the prediction target.

**Figure 1:** Average 5-Fold Cross-Validated Model Performance (bars), +/- 1 Standard Deviation of Cross-Validated Performance (intervals), and Test Set Performance (dots) of Traditional Geospatial Interpolation Compared to GeoFM Embeddings, for Rate-Type Prediction Targets

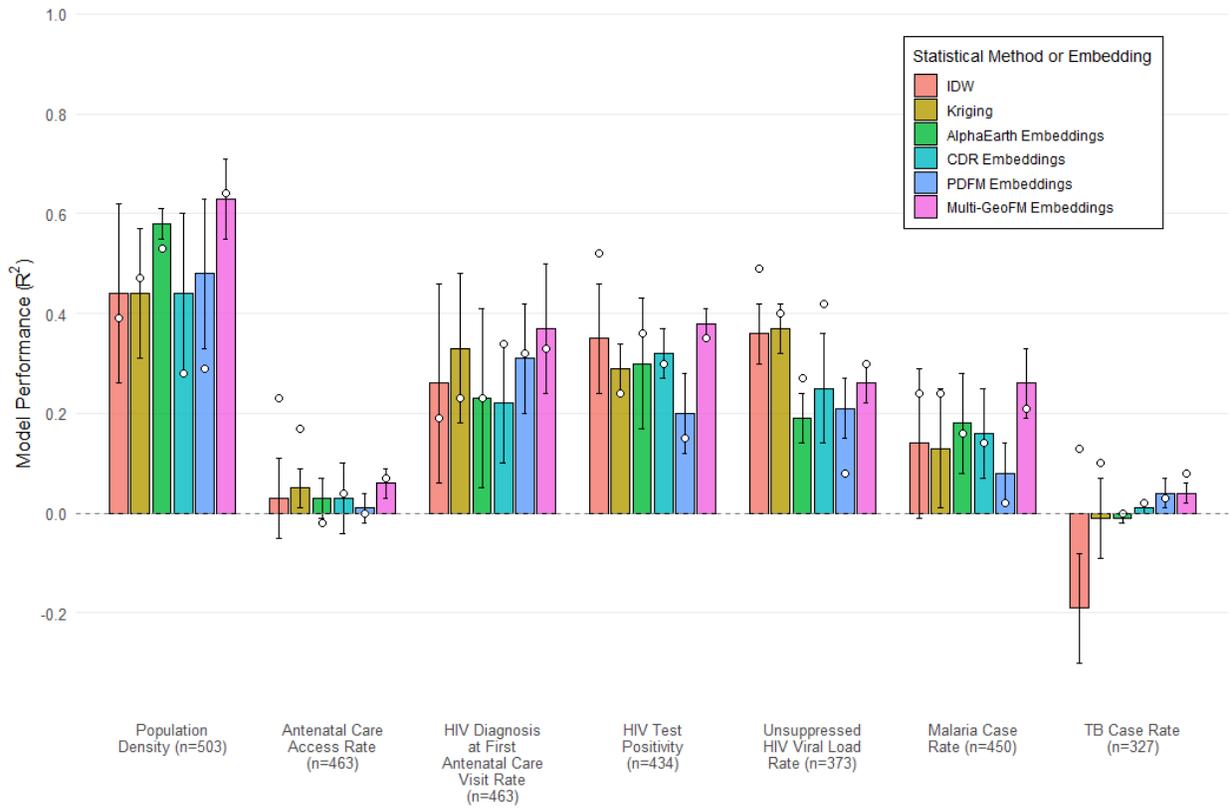

**Figure 2:** Average 5-Fold Cross-Validated Model Performance (bars), +/- 1 Standard Deviation of Cross-Validated Performance (intervals), and Test Set Performance (dots) of Traditional Geospatial Interpolation Compared to GeoFM Embeddings, for Count-Type Prediction Targets

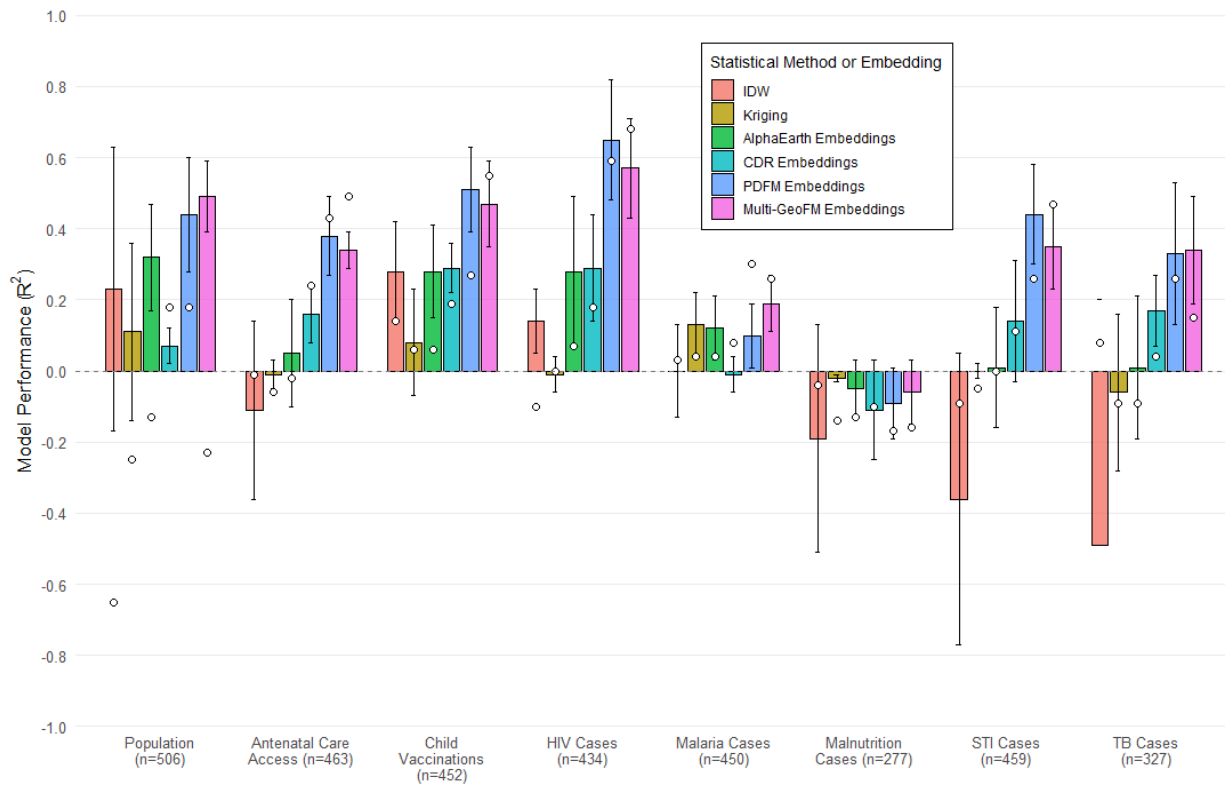

**Baseline Models**

IDW and Kriging provided modest predictive performance, with average R² values of 0.090 and 0.097 from 5-fold cross-validation and the test set, respectively. The main exception was population density, for which IDW had R² of 0.44 and 0.39 from 5-fold cross-validation and the test set, respectively. Kriging yielded R² values of 0.44 and 0.47 for population density from 5-fold cross-validation and the test set, respectively. Unsuppressed HIV viral load rate also demonstrated stronger predictions using IDW with R² values of 0.36 and 0.49 from 5-fold cross-validation and the test set, respectively. Kriging gave R² values of 0.37 and 0.40 from 5-fold cross-validation and the test set, respectively, which exceeded all embeddings-based performance .

**Embedding-based Models**

All embedding-based models performed well when predicting population density, with cross-validated R² values ranging from 0.44 to 0.63 and test set R² values ranging from 0.28 to 0.64. The PDFM embedding-based model provided an R² of 0.51 from cross-validation and 0.27 from the test set, outperforming all other single embeddings models. The PDFM embedding-based model outperformed IDW, Kriging, and all other single embeddings models, population, antenatal care access, HIV cases, STI cases, and TB cases. The AlphaEarth embedding-based model performed best overall for population density and malaria case rate, with R² values of 0.58 and 0.18 from cross-validation, respectively while test set R² values were 0.53 and 0.16, respectively. The CDR embedding-based model did not outperform other embedding-based models but did outperform IDW and Kriging for population density, malaria case rate, TB case rate, antenatal care access, child vaccinations, HIV cases, STI cases, and TB cases based on cross-validated R².

**Multi-GeoFM Model**

The Multi-GeoFM model that included all embedding sources as predictors improved performance for many indicators. Notably, the Multi-GeoFM model achieved the strongest predictions for antenatal care access rate (R²=0.06), malaria cases (R²=0.19), malaria rate (R²=0.26), HIV test positivity (R²=0.38), population density (R²=0.63), population (R²=0.49), TB cases (R²=0.34), TB case rate (R²=0.04), and HIV diagnosis at first antenatal care visit (R²=0.37) based on cross-validated R². When assessed on the test set R², the Multi-GeoFM model performed best on antenatal care access (R²=0.49), child vaccinations (R²=0.55), HIV cases (R²=0.68), population density (R²=0.64), and STI cases (R²=0.47).

IDW, Kriging, and all embeddings-based methods had consistently poor performance for antenatal care access rate, TB case rate, and malnutrition cases.

## Discussion

We evaluated the performance of multiple embedding-based models for predicting health indicators across Malawi, demonstrating that embeddings produced from GeoFMs can meaningfully improve prediction of select outcomes in a low-resource setting. Our findings suggest that while predictive performance varied considerably by health indicator, embedding-based approaches generally outperformed traditional geostatistical methods. Importantly, the Multi-GeoFM model that integrated multiple embedding sources consistently produced the strongest results, underscoring the value of leveraging novel prediction methods.

Population density emerged as the most accurately predicted indicator across all methods. This finding is intuitive, as population distribution is directly observable from satellite imagery and strongly correlated with spatial features such as land use and building structure (13). As a result of this, population density serves as a useful benchmark for evaluating performance across more complex health indicators.

The embedding-based models performed well for several key health outcomes. PDFM embeddings performed especially well for count-based prediction targets including population, antenatal care access, HIV cases, STI cases, and TB cases. It is also worth noting that for these count-type prediction targets, the prediction task is essentially a blend of population and the health output in question, because the values are not normalized to population size. Despite this, PDFM still significantly outperformed IDW and Kriging, implying there is additional value from embeddings based on internet search, maps, and environment.

AlphaEarth satellite embeddings outperformed IDW and Kriging and other single embedding methods for population density and malaria case rate. These results suggest that the satellite-based embedding captured meaningful patterns in human built environment and environmental conditions. For population density, this likely includes visual aspects of building density including roads and structures. For malaria - which is a vector borne disease - the satellite AlphaEarth embeddings are likely finding signals in important seasonal and environmental features including standing water, rivers and lakes, and greenery which are important to the mosquito's habitat.

TB and malnutrition indicators demonstrated weak predictive performance, with $R^2$ values close to zero across all approaches. These outcomes also had the smallest sample sizes contributing usable data after quality control, which likely reflects both reporting limitations and underlying measurement challenges. The limited data available may explain why embeddings offered little improvement for these indicators. Improving the availability and completeness of TB and malnutrition reporting remains critical for future work.

Our results highlight the promise of Multi-GeoFM approaches. No single embedding source consistently outperformed the others, but their integration produced the most robust predictions across a range of indicators. Future research should explore optimal methods for combining embeddings and examine whether additional sources, such as climate or socioeconomic data, further enhance predictive power.

Additionally, advancements in interoperability are facilitating access to rich imagery, population data, and environmental information. In October 2025 Google announced its "Earth AI" family of geospatial AI models (27). Earth AI is also enhanced with a geospatial reasoning agent to seamlessly orchestrate, integrate, and analyze these data for rapid insights using plain language queries and minimal technical ability.

This study builds the case that embeddings can enhance health system monitoring in LMICs, where traditional health data systems are often constrained by incomplete reporting and variable data quality. By integrating nontraditional data streams such as satellite imagery, mobile phone records, and search data, embeddings offer an efficient means of supplementing existing health information systems. While they cannot replace robust primary data collection, embeddings may provide a valuable tool for identifying trends, filling geographic gaps, and informing resource allocation in real time.

**Limitations**
This study has limitations that warrant consideration. First, the analysis was restricted to data from Malawi, and findings may not be directly generalizable to other LMICs with different health system structures, epidemiologic profiles, or data reporting practices. Second, the quality of the prediction targets depended on routine reporting through DHIS2 and LIMS, both of which are known to have variable completeness and accuracy. Indicators with the least available data, such as TB and malnutrition, performed poorly across all models, underscoring the dependence of embedding-based approaches on a minimum amount of underlying data. Third, the exclusion of facilities with high proportions of missing or zero values, while necessary for data quality, may have introduced bias if excluded facilities systematically differed from those retained. Finally for prediction targets including STIs cases, childhood vaccinations, and malnutrition cases - we do not have access to the appropriate population size estimates (sexual health clinic attendance for STIs, number of children under age 5 for vaccinations and malnutrition) to calculate rates, but have included these prediction targets due to their epidemiological importance in Malawi.

**Conclusion**

We found that GeoFM embeddings can have a modest predictive capability in LMIC settings for health and demographics. Results were generally mixed, with the Multi-GeoFM embeddings model approach, demonstrating an additive effect to combining more data sources. Future research should focus on better understanding what types of health conditions are best predicted by various embedding types (i.e., satellite, internet search, phone, social media), the role of temporality, and other prediction tasks such as forecasting and outbreak detection.

# References

1 AbouZahr C, Boerma T. Health information systems: the foundations of public health. Bull World Health Organ. 2005 Aug;83(8):578-83. Epub 2005 Sep 22. PMID: 16184276; PMCID: PMC2626318.

2 Borges do Nascimento IJ, Marcolino MS, Abdulazeem HM, Weerasekara I, Azzopardi-Muscat N, Gonçalves MA, Novillo-Ortiz D. Impact of Big Data Analytics on People's Health: Overview of Systematic Reviews and Recommendations for Future Studies. J Med Internet Res. 2021 Apr 13;23(4):e27275. doi: 10.2196/27275. PMID: 33847586; PMCID: PMC8080139.

3 Syed R, Eden R, Makasi T, Chukwudi I, Mamudu A, Kamalpour M, Kapugama Geeganage D, Sadeghianasl S, Leemans SJJ, Goel K, Andrews R, Wynn MT, Ter Hofstede A, Myers T. Digital Health Data Quality Issues: Systematic Review. J Med Internet Res. 2023 Mar 31;25:e42615. doi: 10.2196/42615. PMID: 37000497; PMCID: PMC10131725.

4 Orlu GU, Abdullah RB, Zaremohzzabieh Z, Jusoh YY, Asadi S, Qasem YAM, Nor RNH, Mohd Nasir WMHb. A Systematic Review of Literature on Sustaining Decision-Making in Healthcare Organizations Amid Imperfect Information in the Big Data Era. Sustainability. 2023; 15(21):15476. https://doi.org/10.3390/su152115476.

5 Charles-Smith LE, Reynolds TL, Cameron MA, Conway M, Lau EH, Olsen JM, Pavlin JA, Shigematsu M, Streichert LC, Suda KJ, Corley CD. Using Social Media for Actionable Disease Surveillance and Outbreak Management: A Systematic Literature Review. PLoS One. 2015 Oct 5;10(10):e0139701. doi: 10.1371/journal.pone.0139701. PMID: 26437454; PMCID: PMC4593536.

6 Bolt K, Gil-González D, Oliver N. Unconventional data, unprecedented insights: leveraging non-traditional data during a pandemic. Front Public Health. 2024 Mar 7;12:1350743. doi: 10.3389/fpubh.2024.1350743. PMID: 38566798; PMCID: PMC10986850.

7 Dritsas E, Trigka M. Remote Sensing and Geospatial Analysis in the Big Data Era: A Survey. Remote Sensing. 2025; 17(3):550. https://doi.org/10.3390/rs17030550

8 Wesolowski A, Buckee CO, Engø-Monsen K, Metcalf CJE. Connecting Mobility to Infectious Diseases: The Promise and Limits of Mobile Phone Data. J Infect Dis. 2016 Dec 1;214(suppl_4):S414-S420. doi: 10.1093/infdis/jiw273. PMID: 28830104; PMCID: PMC5144902.

9 Sun H, Zhang Y, Gao G, Wu D. Internet search data with spatiotemporal analysis in infectious disease surveillance: Challenges and perspectives. Front Public Health. 2022 Dec 5;10:958835. doi: 10.3389/fpubh.2022.958835. PMID: 36544794; PMCID: PMC9760721.

10 Vatsavai RR. Geospatial Foundation Models: Recent Advances and Applications. BigSpatial '24: Proceedings of the 12th ACM SIGSPATIAL International Workshop on Analytics for Big Geospatial Data. 2024 Oct 29. doi: https://doi.org/10.1145/3681763.3698478.

11 Haw DJ, Pung R, Read JM, Riley S. Strong spatial embedding of social networks generates nonstandard epidemic dynamics independent of degree distribution and clustering. Proc Natl Acad Sci U S A. 2020 Sep 22;117(38):23636-23642. doi: 10.1073/pnas.1910181117. Epub 2020 Sep 8. PMID: 32900923; PMCID: PMC7519285.

12 Agarwal M, Sun M, Kamath C, Muslim A, Sarker P, Paul J, Yee H, Sieniek M, Jablonski K, Mayer Y, Fork D, de Guia S, McPike J, Boulanger A, Shekel T, Schottlander D, Xiao Y, Manukonda MC, Liu Y, Bulut N, Abu-el-haija S, Eigenwillig A, Kothari P, Perozzi B, Bharel M, Nguyen V, Barrington L, Efron N, Matias Y, Corrado G, Eswaran K, Prabhakara S, Shetty S, Prasad G. General Geospatial Inference with a Population Dynamics Foundation Model. Arxiv. 2024 Dec 28. arXiv:2411.07207v3.

13 Klemmer K, Rolf E, Robinson C, Mackey L, Rußwurm M. Proceedings of the AAAI Conference on Artificial Intelligence. 2025 Apr 11. doi: https://doi.org/10.1609/aaai.v39i4.32457

14 Szwarcman D, Roy S, Fraccaro P, Gislason PE, Blumenstiel B, Ghosal R, Henrique de Oliveira P, Lucas de Sousa Almeida J, Sedona R, Kang Y, Chakraborty S, Wang S, Gomes C, Kumar A, Truong M, Godwin D, Lee H, Hsu CY, Asanjan AA, Mujeci B, Shidham D, Keenan T, Arevalo P, Li W, Alemohammad H, Olofsson P, Hain C, Kennedy R, Zadrozny B, Bell D, Cavallaro G, Watson C, Maskey M, Ramachandran R, Moreno JB. Prithvi-EO-2.0: A Versatile Multi-Temporal Foundation Model for Earth Observation Applications. 2025 Feb. arXiv arXiv:2412.02732. doi: https://doi.org/10.48550/arXiv.2412.02732

15 Tseng G, Cartuyvels R, Zvonkov I, Purohit M, Rolnic D, Kerner H. Lightweight, Pre-Trained Transformers for Remote Sensing Timeseries. 2024 Feb. arXiv arXiv:2304.14065. doi: https://doi.org/10.48550/arXiv.2304.14065

16 Clay Foundation Model https://clay-foundation.github.io/model/index.html

17 Feng Z, Atzberger C, Jaffer S, Knezevic J, Sormunen S, Young R, Lisaius MC, Immitzer M, Jackson T, Ball J, Coomes DA, Madhavapeddy A, Blake A, Keshav S. TESSERA: Precomputed FAIR Global Pixel Embeddings for Earth Representation and Analysis. 2025 Jun. arXiv arXiv:2506.20380 doi: https://doi.org/10.48550/arXiv.2506.20380

18 Tseng G, Fuller A, Riel M, Herzog H, Beukema P, Bastani F, Green JR, Shelhamer E, Kerner H, Rolnick D. Galileo: Learning Global & Local Features of Many Remote Sensing Modalities. 2025 Jun. arXiv arXiv:2502.09356 doi: https://doi.org/10.48550/arXiv.2502.09356

19 Carvalho RMS, Oliveira D, Pesquita C. Knowledge Graph Embeddings for ICU readmission prediction. BMC Med Inform Decis Mak. 2023 Jan 19;23(1):12. doi: 10.1186/s12911-022-02070-7. PMID: 36658526; PMCID: PMC9850812.

20 Hoxha K, Hung YW, Irwin BR, Grépin KA. Understanding the challenges associated with the use of data from routine health information systems in low- and middle-income countries: A systematic review. Health Inf Manag. 2022 Sep;51(3):135-148. doi: 10.1177/1833358320928729. Epub 2020 Jun 30. PMID: 32602368.

21 Kemp S. Digital 2020: Malawi.2020 Feb 18. https://datareportal.com/reports/digital-2020-malawi

22 World Health Organization. Malawi. 2023. https://data.who.int/countries/454

23 World Health Organization. United States of America. 2023. https://data.who.int/countries/840

24 Krugman A. The State of Global Health Funding. Think Global Health. 2025 August. https://www.thinkglobalhealth.org/article/state-global-health-funding-august-2025

25 Brown CF, Kazmierski MR, Pasquarella VJ, Rucklidge WJ, Samsikova M, Zhang C, Shelhamer E, Lahera E, Wiles O, Ilyushchenko S, Gorelick N, Zhang LL, Alj S, Schechter E, Askay S, Guinan O, Moore R, Boukouvalas A, Kohli P (2025. AlphaEarth Foundations: An embedding field model for accurate and efficient global mapping from sparse label data. arXiv arXiv.2507.22291. doi:10.48550/arXiv.2507.22291

26 Moszczynski M, Asbah S, Metz L, Haggard R, Cooper H, Smith T, Green D. (In Preparation). Generating General Purpose Embeddings using a Graphical Neural Network on Anonymized Mobile Phone Call Detail Records in Malawi. 2025 Oct.

27 Bell A, Aides A, Helmy A, Muslim A, Barzilai A, Slobodkin A, Jaber B, Schottlander D, Leifman G, Paul J, Sun M, Sherman N, Williams N, Bjornsson P, Lee R, Alcantara R, Turnbull T, Shekel T, Silverman V, Gigi Y, Boulanger A, Ottenwess A, Ahmadalipour A, Carter A, Elliott C, Andre D, Aharoni E, Jung G, Thurston H, Bien J, McPike J, Rothenberg J, Hegde K, Markert K, Jablonski KP, Houriez L, Bharel M, VanLee P, Sayag R, Pilarski S, Cazares S, Pasternak S, Jiang S, Jiang S, Colthurst T, Chen Y, Refael Y, Blau Y, Carny Y, Maguire Y, Hassidim A, Manyika J, Thelin T, Beryozkin G, Prasad G, Barrington L, Matias Y, Efron N, Shetty S. Earth AI: Unlocking Geospatial Insights with Foundation Models and Cross-Modal Reasoning. 2025 Oct. arXiv arXiv:2510.18318 doi: https://doi.org/10.48550/arXiv.2510.18318